\documentclass[%
 reprint,
nofootinbib,
bibnotes,
 amsmath,amssymb,
 aps,
 prl
]{revtex4-2}
\usepackage{graphicx}
\usepackage{dcolumn}
\usepackage{bm}
\usepackage{sidecap}
\usepackage{subfigure}
\usepackage{mathtools}
\usepackage{multirow}
\usepackage{booktabs}
\usepackage{makecell}
\usepackage{colortbl}
\usepackage{float}
\usepackage{multibib}
\usepackage[dvipsnames]{xcolor}
\usepackage[colorlinks,
linkcolor=NavyBlue,
anchorcolor=blue,
urlcolor=NavyBlue,
citecolor=NavyBlue
]{hyperref}
\usepackage{marvosym}
\usepackage{appendix}
\setcitestyle{numbers,super}

\makeatletter 
\renewcommand\onecolumngrid{
	\do@columngrid{one}{\@ne}%
	\def\set@footnotewidth{\onecolumngrid}
	\def\footnoterule{\kern-6pt\hrule width 1.5in\kern6pt}%
}

\renewcommand\twocolumngrid{
	\def\footnoterule{
		\dimen@\skip\footins\divide\dimen@\thr@@
		\kern-\dimen@\hrule width.5in\kern\dimen@}
	\do@columngrid{mlt}{\tw@}
}%
\makeatother  
\renewcommand{\figurename}{\textbf{Fig.}}

\makeatletter
\renewcommand*{\fnum@figure}{{\normalfont\bfseries\figurename~\thefigure}}
\renewcommand*{\@caption@fignum@sep}{\textbf{\,$|$\,}}
\makeatother
\newcounter{myfigure}
\makeatletter
\@addtoreset{figure}{myfigure}
\makeatother
\newcommand{\beq}{\begin{equation}}
\newcommand{\eeq}{\end{equation}}
\newcommand{\md}{\mathrm{d}}
\newcommand{\mi}{\mathrm{i}}
\begin{document}

\preprint{APS/123-QED}
\let\thefootnote\relax
\title{Emergence Transformer: Dynamical Temporal Attention Matters}
\author{Zihan Zhou\textsuperscript{1,2,\#}}
\author{Bo-Wei Qin\textsuperscript{1,2,3,4\#,\Letter}}
\author{Kai Du\textsuperscript{1,\Letter}}
\author{Wei Lin\textsuperscript{1,2,3,4\Letter}}
\affiliation{%
${}^1$School of Mathematical Sciences and Shanghai Center for Mathematical Sciences, Fudan University, 200433 Shanghai, China\\
${}^2$Research Institute of Intelligent Complex Systems, Fudan University, 200433 Shanghai, China\\
${}^3$Shanghai Artificial Intelligence Laboratory, 200232 Shanghai, China\\
${}^4$State Key Laboratory of Medical Neurobiology and MOE Frontiers Center for Brain Science, Institute of Brain Science, Fudan University, 200032 Shanghai, China
}%
\begin{abstract}
The Transformer, a breakthrough architecture in artificial intelligence, owes its success to the attention mechanism, which utilizes long-range interactions in sequential data, enabling the emergent coherence between large language models (LLMs) and data distributions. However, temporal attention, that is, different forms of long-range interactions in temporal sequences, has rarely been explored in emergence phenomenon of complex systems including oscillatory coherence in quantum, biophysical, or climate systems. Here, by designing dynamical temporal attention (DTA) with time-varying query, key, and value matrices, we propose an {\em Emergence Transformer}. This architecture allows each component to interact with its own or its neighbors' past states through dynamical attention kernels, thereby enabling the promotion and/or suppression of the emergent coherence of components. Interestingly, we uncover that {\em neighbor-DTA} consistently promotes oscillatory coherence, whereas {\em self-DTA} exhibits an optimal attention weight for coherence enhancement, owing to its non-monotonic dependence on network structure. Practically, we demonstrate how DTA reshapes social coherence, suggesting strategies to either enhance agreement or preserve plurality. We further apply DTA to the paradigmatic Hopfield neural network, achieving emergent continual learning without catastrophic forgetting. Together, these results lay a foundation and provide an immediate paradigm for modulating emergence phenomenon in networked dynamics only using DTA.
\end{abstract}
\maketitle

Coherence-incoherence transition of interacting components is ubiquitous in complex systems. Those transitions often relate to the emergence of new phenomenon and/or functions, among which large language model (LLM) is the most notable example recently. Ordering sequential tokens appropriately as coherent and structured context yields the emergence of representational ability of LLMs. In light of the attention mechanism in the milestone architecture Transformer\cite{Vaswani2017}, the coherence between model prediction and data distribution is further enhanced. In addition to the LLMs\cite{Kirk2024}, the emergence phenomena owing to coherence-incoherence transitions are also prevalent in dynamically-evolving complex systems\cite{Pikovsky2010,Wong2023} ranging from quantum optics\cite{Bloch2022,Moille2023}, systems biology and medicine \cite{Domenico2023}, artificial intelligence\cite{Yan2024} to power and ecological systems\cite{Artime2024,Chang2023}. However, compared to the remarkable success of Transformer in LLMs, a dynamical architecture for mastering temporal sequences of complex systems and their coherent or incoherent states relating to the emergence phenomena is still lacking. Unaddressed and challenging questions are: Can dynamical counterparts of query, key and value matrices as those in the classical Transformer be designed, and if so, how to implement them in the model and what are their roles in manipulating the emergent coherence or incoherence? Here, we address those challenges by developing an {\em Emergence Transformer} framework that incorporates dynamical temporal attention (DTA). Building upon the classical Transformer with static attention\cite{Vaswani2017}, we indeed propose a fundamentally different architecture incorporating dynamically-updating attention. 
	
We explore the role of {\em Emergence Transformer} in the context of synchronization of coupled oscillators, one of the most significant emergence phenomena of complex systems\cite{Pikovsky2010}.~It coordinates interacting components, driving them from an incoherent state toward a coherent one, or vice versa.~Actually, the alignment of LLMs to individuals\cite{Kirk2024} is also a type of emergent coherence (or synchronization). Another example is coherent oscillations as a mechanistic basis for a neural filter\cite{Raccuglia2025}.~In many practical scenarios, how coherence/incoherence emerges (or the level of synchronization) plays a significant role in maintaining systems' functionality.~In LLMs, for instance, a moderate alignment favors generalization, whereas domain-specific models need to be fine-tuned to strong coherence\cite{Zhou2023}.~Similarly, in oscillatory complex systems, synchronization plays a profoundly dual role\cite{Osipov2007,Sugitani2021}.~On the one hand, coherence is essential for sustaining proper functions.~It optimizes quantum clock networks\cite{Komar2014}, stabilizes power grids\cite{Rohden2012,Smith2022}, and underpins the associative memory retrieval of Hopfield neural networks (HNNs)\cite{Min2025,Fu2025}.~On the other hand, excessive coherence is detrimental, exacerbating locust outbreaks\cite{Liu2024}, triggering pathological seizures\cite{Ren2021}, and disrupting physiological transitions such as sleep-wake cycle\cite{diSanto2018}. The inherent duality of the emergent coherence presents a fundamental challenge for its flexible modulation.

Previous works mainly focused on two factors of oscillatory systems: the distribution of oscillators' natural frequencies\cite{Kuramoto1975} and the network topology\cite{Zhang2024,Millan2025}.~Canonical models are the coupled Stuart-Landau oscillators\cite{Zhang2021a} (or the Hopf normal form\cite{Nijholt2022,Zhong2023}), and the celebrated Kuramoto phase model\cite{Kuramoto1975,Acebron2005}. Based on these models, analytical methods including the mean-field approximation\cite{Strogatz2000,Medvedev2014,Gkogkas2022} [see also Supplementary Information (SI) Sec.~1] and the Ott-Antonsen ansatz\cite{Ott2008,Omelchenko2022} have been developed. They provided pioneering insights into how network topology and natural frequencies shape coherence and synchronizability\cite{Motter2013,Nazerian2024,Lee2025,Buendia2025}.~The former one is the level of synchronization, and the latter one represents the propensity of emergence. However, both factors are often intrinsic and static, offering little room for modulation. Contrarily, we show that the DTA in the {\it Emergence Transformer} is more flexible and has a dual role in promoting and/or suppressing emergent coherence.

The core of emergent coherence is appropriate information exchange.~It generally occurs instantaneously and spatially, driving components to synchronize with their neighbors or external information.~For instance, the LLMs receive information from sequential data and achieve high coherence by attention, a mechanism leveraging non-local spatial information\cite{Vaswani2017}.~However, utilizing appropriately the information of dynamically-updating temporal sequence was often overlooked in previous studies. Given that every component in a complex system has a temporal sequence, a natural question arises: What if a component also interact with its own and even its neighbors' past states in a well-designed structure?

Actually, recent studies demonstrated that incorporating temporal information in reservoir computing significantly reduces the spatially extensive inputs, and enables coherence between a reservoir and a target system with only a few nodes\cite{Appeltant2011,Duan2023}. Therefore, utilizing temporal information provides a potential approach for modulating the emergent coherence.~Also, a previous work showed that a single oscillator interacting with its own temporal sequence exhibits emergent coherence with its past states even with the presence of noise\cite{Benaim2002}.~More strikingly, the stationary distribution of states along this single, time-evolving trajectory is mathematically equivalent to the snapshot distribution of states in a globally coupled ensemble of infinitely many oscillators, see SI Sec.~2 and Fig.~S1 for more details. This remarkable equivalence suggests that the temporal information compensates those distributed ones in space.~Those insights guide us to design appropriately the structure of DTA in the {\em Emergence Transformer} for flexibly modulating emergent coherence, more specifically, in promoting and/or suppressing synchronization by different attention kernels.\\
\newline
\noindent{\bf Dynamics of the {\em Emergence Transformer} }\\

\begin{figure*}[!hp]
	\subfigure{\label{Fig1:a}}
	\subfigure{\label{Fig1:b}}
	\subfigure{\label{Fig1:c}}
	\subfigure{\label{Fig1:d}}
	\subfigure{\label{Fig1:e}}
	\subfigure{\label{Fig1:f}}
	\subfigure{\label{Fig1:g}}
	\includegraphics[scale=1]{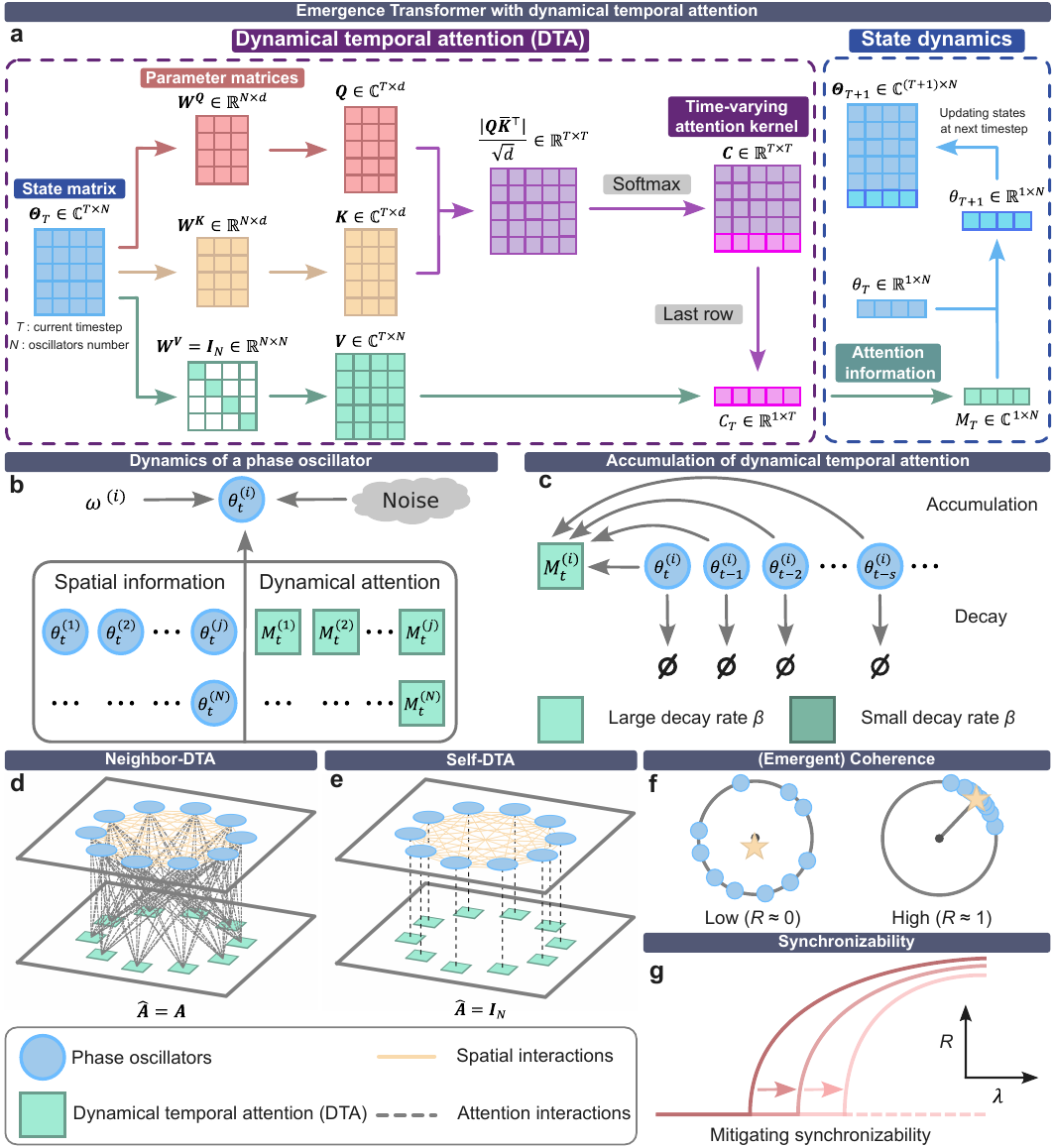}
	\caption{\label{Fig1} {\bf {\em Emergence Transformer} and emergent coherence.} {\bf a} Illustration of phase, query, key and value matrices and the attention kernel together with the attention information used for updating the phase dynamics. {\bf b}, The dynamics of a phase oscillator $\theta^{(i)}$ in {\em Emergence Transformer} is affected by the current states (blue circles) and attention information (green squares) of its neighbors (depending on the network topology), its own natural frequency $\omega^{(i)}$ and noise. {\bf c}, A natural scenario of the accumulation of the attention information of an oscillator. It is accumulated from all of its past states (blue circles) with an exponential decay rate $\beta$. Light and dark green represent the attention information with large and small decay rate, respectively. The schematic diagram shows a discrete-time case just for the sake of better understanding. {\bf d}, The {\em Emergence Transformer} with the {\it neighbor-DTA}. {\bf e}, The {\em Emergence Transformer} with the {\it self-DTA}. {\bf f}, The coherence characterizes the level of synchronization by using the classical order parameter $R$. {\bf g}, Synchronizability refers to the propensity to synchronization. Larger critical coupling strength $\lambda_c$ corresponds to lower synchronizability.	}

\end{figure*}

We introduce the {\em Emergence Transformer} by considering the dynamics of the phase oscillators and elucidate how we incorporate DTA. A general model of $N$ phase oscillators reads (Figs.~\ref{Fig1:a}--\ref{Fig1:c})
\begin{equation}\label{eq:GeneralKuramoto}
	\dot{\theta}_t^{(i)}=\omega^{(i)}+\lambda{\rm Im}[\mathcal{I}^{(i)}_t{\rm e}^{-{\rm i}\theta^{(i)}_t}]+\xi^{(i)}_t,~~~i=1,2,\dots,N,
\end{equation}
where $\theta^{(i)}_t$ is the phase of the $i$th oscillator at time $t$, $\omega^{(i)}$ is its natural frequency assigned randomly from a given distribution $g(\omega)$. This distribution often satisfies certain conditions\cite{Acebron2005}. In this work, it is assumed to be uni-modal and symmetric to the average $\mathbb{E}[\omega]=0$. The oscillators are coupled through the second term on the right-hand side, where ${\rm Im}[\cdot]$ is the imaginary part of the argument and $\mathcal{I}^{(i)}_t$ represents the total received information. As we will see, it includes both dynamically-temporal and instantaneously-spatial information. Moreover, $\lambda$ is the global coupling strength, and $\xi_t^{(i)}$ is the Gaussian white noise satisfying $\mathbb{E}[\xi^{(i)}_t]=0$, $\mathbb{E}[\xi^{(i)}_t\xi^{(j)}_s]=2D\delta_{ij}\delta(t-s)$. Here, $\delta_{ij}$ and $\delta(\cdot)$ are the Kronecker delta and Dirac delta function, respectively, and $D$ represents the noise strength.

Regarding the total information received by the $i$th oscillator, it is either spatial, attention or their combination. Specifically, here we have
\begin{equation}\label{eq:TotalInformation}
	\mathcal{I}^{(i)}_t=(1-\alpha)\frac{1}{d_i}\sum_{j=1}^{N}A_{ij}{\rm e}^{{\rm i}\theta^{(j)}_t}+\alpha\frac{1}{\hat{d}_i}\sum_{j=1}^{N}\hat{A}_{ij}M_t^{(j)},
\end{equation}
where ${\rm e}^{{\rm i}\theta_t^{(j)}}$ and $M_t^{(j)}$ are, respectively, the spatial and attention information (Fig.~\ref{Fig1:b}) contributed by the $j$th oscillator at time $t$. The attention weight is characterized by the parameter $\alpha\in[0,1]$. 

Each attention information $M_t^{(j)}$ actually stores the non-local information of the $j$th oscillator's temporal sequence based on certain attention kernel. It can be derived by using suitable and trainable parameter matrices ${\bm W}^{{\bm Q}}\in\mathbb{R}^{N\times d}$, ${\bm W}^{{\bm K}}\in\mathbb{R}^{N\times d}$ and ${\bm W}^{{\bm V}}\in\mathbb{R}^{N\times N}$ as those in the Transformer architecture\cite{Vaswani2017}, where $d$ is the number of features of the temporal sequence. For a better understanding, we illustrate a discrete-time case in Fig.~\ref{Fig1:a}. Its detailed derivation is given in Methods. 

As for the continuous-time phase oscillator, we define ${\bm \Theta}_t(s)\coloneqq\left[{\rm e}^{{\rm i}\theta_s^{(1)}},{\rm e}^{{\rm i}\theta_s^{(2)}},\dots,{\rm e}^{{\rm i}\theta_s^{(N)}}\right]$, which is a vector-valued function from $[0,t]$ to $\mathbb{C}^N$.~Note that $t$ is the current time and changes continuously as time evolves.~Thus, ${\bm \Theta}_t$ stores the temporal sequence of every oscillator. Then, the query and key matrices are computed respectively as ${\bm Q}(s)={\bm \Theta}_t(s){\bm W}^{\bm Q}$ and ${\bm K}(s)={\bm \Theta}_t(s){\bm W}^{\bm K}$.~The two matrices are then used to compute the attention kernel. Specifically, a two-variable function $U(s,\tau)=\|{\bm Q(s)}\bar{\bm K}(\tau)\|/\sqrt{d}$ is obtained.~It represents the un-normalized attention kernel from time $s$ to $\tau$.~The conjugate operation is applied to one of the matrices so that the attention weight is maximized when the two states are parallel.~For each time $s$, a softmax operator is then applied to obtain normalized attention kernel as $C_s(\tau)={\rm e}^{U(s,\tau)}/\int^t_0{\rm e}^{U(s,\tau)}\md\tau$.~It delineates the relative attention from time $s$ to other time $\tau$. Here, we assume that the attention is solely paid to the past states, i.e., $\tau<s$. Therefore, we only use the function $C_t(\tau)$ at the current time $t$ which is analogous to using the last row of the attention kernel for the discrete-time case (Fig.~\ref{Fig1:a}). As we want to compute the attention information for each oscillator independently, ${\bm W}^{\bm V}$ is set to the identity matrix ${\bm I}_N$.~Thus, the value matrix containing the phase states becomes ${\bm V(s)}={\bm \Theta}_t(s){\bm W}^{\bm V}={\bm \Theta}_t(s)$. Finally, the attention information in Eq.~\eqref{eq:TotalInformation} is computed as ${\bm M}_t=\int^t_0C_t(\tau){\bm \Theta}_t(\tau)\md\tau$ implying that
\beq\label{eq:AttentionInformation}
M_t^{(i)}=\int^t_0C_t(\tau){\rm e}^{\mi\theta_\tau^{(i)}}\md\tau,~~~i=1,2,\dots,N.
\eeq
We emphasize again that the final attention kernel $C_t(\tau)$ changes as time $t$ evolves.~Therefore, in our {\em Emergence Transformer}, the introduced DTA is different from static attentions or their linear/nonlinear combinations.\\
\newline
{\bf A natural dynamics of DTA}\\

From Eq.~\eqref{eq:AttentionInformation}, the attention information $M^{(i)}_t$ is actually the accumulation of past states of the $i$th oscillator. The accumulation rule is characterized by the attention kernel $C_t(\tau)$ depending on the parameter matrices ${\bm W}^{\bm Q}$ and ${\bm W}^{\bm K}$.~Generally, both matrices (or attention kernel) are pre-designed according to certain mechanisms or learned for achieving desired outcomes. Note that the accumulation of attention is, to some extent, analogous to memorize the past states. We therefore consider here not only the accumulation but also a natural decaying of the attention information with rate $\beta$ (Fig.~\ref{Fig1:c}). Indeed, such a decaying mechanism is a ubiquitous dynamical process in nature especially for human memory\cite{Candia2019}.~Thus, we consider the dynamics of attention information as
\begin{equation}\label{eq:MemoryDynamics}
	\dot{M}_t^{(i)}=\beta\left[{\rm e}^{{\rm i}\theta^{(i)}_t}-{M_t^{(i)}}\right],~~i=1,2,\dots,N.
\end{equation}
Consequently, an oscillator pays more attention on its past state at a time that is closer to the present. The decaying rate $\beta$ indeed tunes the structure of the attention kernel. Now, Eqs.~\eqref{eq:GeneralKuramoto}--\eqref{eq:MemoryDynamics} describe the entire dynamics of the proposed {\em Emergence Transformer} for this study.~Solving Eq.~\eqref{eq:MemoryDynamics} explicitly, the attention is in the form of Eq.~\eqref{eq:AttentionInformation} and the entire dynamics of the {\em Emergence Transformer} is a set of delay differential equations (DDEs).\\
\newline
{\bf Spatial and attention networks}\\

Let us now introduce the two networks for transmitting the spatial and the dynamical attention information, i.e., ${\rm e}^{{\rm i}\theta_t^{(i)}}$ and $M_t^{(i)}$, respectively.~The interacting matrices are ${\bm A}=\{A_{ij}\}_{i,j=1}^N$ and $\hat{\bm A}=\{\hat{A}_{ij}\}_{i,j=1}^N$ for spatial and attention networks, respectively.~We assume that both matrices are static and symmetric in this work.~The element $A_{ij}=1$ when the $i$th and the $j$th oscillators transmitting their spatial information, otherwise, $A_{ij}=0$.~The element $\hat{A}_{ij}$ follows the same configuration based on the transmission of the attention information.~Parameters $d_i$ and $\hat{d}_i$ are degrees of the $i$th oscillator in the spatial and attention networks, respectively. They average the received spatial and dynamical attention information separately. We remark that when there is no attention ($\alpha=0$), Eq.~\eqref{eq:GeneralKuramoto} is equivalent to the classical Kuramoto model with noise.

As we will see soon, the above configuration allows us to analyze how attention affects the level and the onset of synchronization (i.e., coherence and synchronizability). Especially, we focus on the effect of attention weight $\alpha$ for different cases with various network topologies. In each case, the networks are fixed once they are configured. For the spatial network, we mainly focus on the fully-connected and the Watts-Strogatz\cite{Watts1998,Strogatz2001} (WS) ones with relatively low rewiring probability.~Other representative and real-world topologies will also be considered when verifying the conclusion.~As for the topology of the attention network, we consider two scenarios.~For the first one, we assume that $\hat{\bm A}={\bm A}$ which means that the transmission of the spatial and the dynamical attention information share the same network.~This refers to a practical scenario when we are allowed to use the originally-established network to transmit the dynamical attention information across neighbors (Fig.~\ref{Fig1:d}).~If this is not the case, we only leverage each oscillator's own attention information (Fig.~\ref{Fig1:e}).~Specifically, the dynamical attention information is used locally by setting $\hat{\bm A}={\bm I}_N$ (identity matrix).~We refer to the former and latter scenarios as {\em Emergence Transformer} with {\em neighbor-DTA} and {\em self-DTA}, respectively.~The detailed models are provided in SI~Sec.~S.4. \\
\newline
\noindent{\bf DTA modulating emergent coherence}\\

As mentioned before, we are interested in two properties of synchronization: coherence and synchronizability. To assess the former one, we use the classical order parameter $R=\left\lvert N^{-1}{\sum_{j=1}^N\rm e}^{{\rm i}\theta_t^{(j)}}\right\rvert\in[0,1]$ to quantify the level of synchronization.~Less $R$ represents lower coherence (Fig.~\ref{Fig1:f}).~As for the latter one, we use the value of critical coupling strength $\lambda_c$ to describe propensity of the emergence of synchronization (Fig.~\ref{Fig1:g}). 
\begin{figure*}[!t]
	\subfigure{\label{Fig2:a}}
	\subfigure{\label{Fig2:b}}
	\subfigure{\label{Fig2:c}}
	\subfigure{\label{Fig2:d}}
	\subfigure{\label{Fig2:e}}
	\subfigure{\label{Fig2:f}}
	\includegraphics[width=17.8cm]{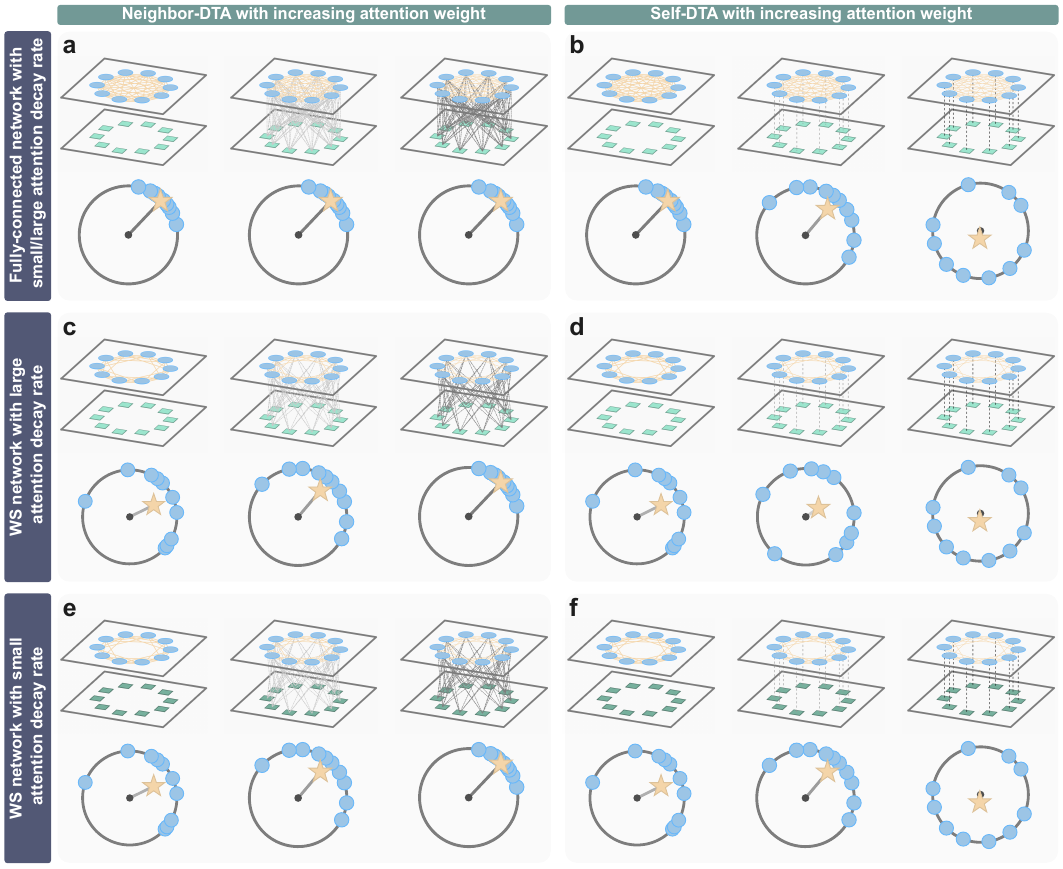}
	\caption{\label{Fig2} {\bf Overview pictures of how DTA affects the coherence.} In all schematic diagrams, we use ten individuals for clarity. The results are actually obtained for populations with more individuals (see Methods for parameters used). Blue dots represent the snapshot of phase oscillators after a sufficiently long time period and yellow pentagram indicates the order parameter (coherence). {\bf a}, Fully-connected (spatial) network with the {\it neighbor-DTA}. As attention weight $\alpha$ increases, the coherence keeps static. {\bf b}, Fully-connected network with the {\it self-DTA}. Increasing $\alpha$ suppresses the coherence. {\bf c}, The WS network with the {\it neighbor-DTA} when attention decay rate is large ($\beta=1$). Increasing $\alpha$ promotes the coherence. {\bf d}, The same settings as those in panel {\bf c} are applied except that the {\it self-DTA} is used which suppress the coherence. {\bf e}, The WS network with the {\it neighbor-DTA} when attention decay rate is small ($\beta=0.01$). Increasing $\alpha$ promotes the coherence. {\bf f}, The same settings as those in panel {\bf e} are applied except that the {\it self-DTA} are used. Increasing $\alpha$ first promotes and then suppress the coherence which is a non-monotonic behavior.}
\end{figure*}

To see how DTA affects synchronization, we provide an overview picture of the results regarding coherence under a fixed coupling strength $\lambda>\lambda_c$ (Fig.~\ref{Fig2}).~We first consider a fully-connected (spatial) network.~The coupling strength is sufficiently large such that $R\approx1$ when $\alpha=0$ (Figs.~\ref{Fig2:a}--\ref{Fig2:b}).~After incorporating DTA, we find that the coherence remains unchanged when leveraging {\it neighbor-DTA} (Fig.~\ref{Fig2:a}) for all attention weight $\alpha$, whereas the coherence keeps being mitigated as $\alpha$ increases in a {\it self-DTA} case (Fig.~\ref{Fig2:b}).~Moreover, we obtain the same results for both small and large decay rate ($\beta=0.01$ and $1$, respectively). Here are some explanations for the mitigation of coherence. Adding {\it self-DTA} reduces the spatial information that is originally large enough for yielding high coherence. Although each single oscillator could synchronize with its past state\cite{Benaim2002}, the entire population is still incoherent.~The reason is that different oscillator synchronizes towards a distinct phase which is a random variable\cite{Benaim2002}.~This result suggests that we can utilize {\it self-DTA} to suppress the emergent coherence (i.e., desynchronization) for a fully-connected network.

We then consider a spatial network with the WS topology that has a larger average shortest path length (ASPL) compared to the fully-connected one (Extended~Data~Tab.~\ref{ExTab1}).~Therefore, it is less effective for transmitting the spatial information.~Consequently, the coherence is less for the same coupling strength $\lambda$ when $\alpha=0$ (Figs.~\ref{Fig2:c}--\ref{Fig2:f}).~Then, we also add DTA and explore its effect.~For attention with both small and large decay rate, $\beta=0.01$ and $1$, adding the {\it neighbor-DTA} yields higher coherence (Figs.~\ref{Fig2:c}~and~\ref{Fig2:e}). This is because the {\it neighbor-DTA} not only compensates the spatial information, but even provides more information from those past states.~Therefore, leveraging the {\it neighbor-DTA} promotes synchronization. 

The results become more surprising for the {\it self-DTA} as they depend on the decay rate $\beta$.~For a larger one ($\beta=1$), the coherence is always mitigated by the {\it self-DTA} (Fig.~\ref{Fig2:d}) because the attention decays too fast to compensate the spatial information.~As for $\beta=0.01$ that is relatively small, non-monotonicity of coherence occurs (Fig.~\ref{Fig2:f}).~As the attention weight $\alpha$ increases, the coherence is first enhanced and then mitigated to a lower value than the initial one.~The reason is that the ``effective" spatial information is small in a WS network yielding a low initial coherence ($\alpha=0$).~Thus, adding appropriate amount of the {\it self-DTA} provides more information by using past states that promotes the synchronization.~But, when the attention weight $\alpha$ is too large, distinct oscillator cannot synchronize towards the same phase thus suppressing the emergent coherence. This is the same as those happen in a fully-connected network with the {\it self-DTA} (Fig.~\ref{Fig2:b}).~The non-monotonic result indeed provides a flexible way for modulating coherence, especially for networks with large ASPL.~It is worth pointing out that the above results are verified under different coupling strength $\lambda>\lambda_c$. \\
\newline
\noindent{\bf Anticipating emergence with DTA}\\

We have discovered diverse roles of DTA in promoting and/or suppressing synchronization by examining coherence.~Now, we analyze it from the propensity of emergence.~As we will see later, the results obtained from both properties are consistent when the variance of natural frequencies, $g(\omega)$, is small which is the case we focus on in this study. Moreover, we can predict analytically the critical coupling strength $\lambda_c$ by considering the continuum limit $N\rightarrow\infty$ for fully-connected networks. This provides us rigorous evidences to support the findings. 
\begin{figure}[!b]
	\subfigure{\label{Fig3:a}}
	\subfigure{\label{Fig3:b}}
	\subfigure{\label{Fig3:c}}
	\subfigure{\label{Fig3:d}}
	\includegraphics[width=8.8cm]{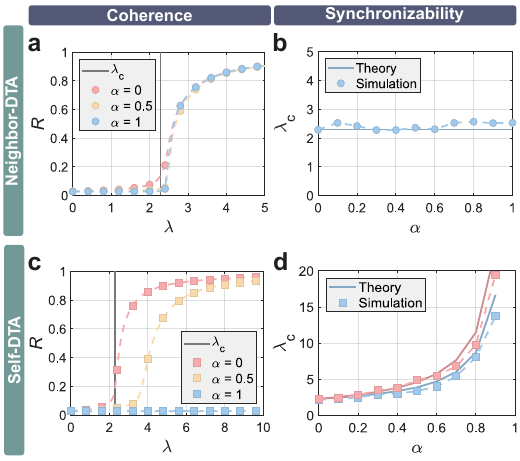}
	\caption{\label{Fig3} {\bf The effect of DTA in fully-connected networks.} Panels {\bf a}--{\bf b} show the results after applying the {\it neighbor-DTA}. Panel {\bf a} presents the order parameter $R$ for different coupling strength $\lambda$ with different attention weight $\alpha$ (red: 0; yellow: 0.5; blue: 1). The vertical line indicates the critical coupling strength $\lambda_c\approx2.28$ obtained analytically. Panel {\bf b} shows the analytical approximation (solid curve) and numerical results (dots) of the critical coupling strength $\lambda_c$ for different $\alpha$. Panels {\bf c}--{\bf d} show the results after applying the {\it self-DTA}. Panel {\bf c} presents the order parameter $R$ for different $\lambda$ and $\alpha$. The vertical line indicates the critical coupling strength $\lambda_c\approx2.28$ obtained analytically when $\alpha=0$. Panel {\bf d} shows the critical coupling strength $\lambda_c$ for different $\alpha$ with distinct attention decay rate (red: $\beta=1$; blue: $\beta=0.01$).}
\end{figure}

We apply the mean-field approximation together with the Fokker-Planck equation\cite{Sonnenschein2012,Zou2023} (FPE) to predict the critical coupling strength $\lambda_c$.~At the continuum limit, we are interested in the conditional phase distribution $\rho(\theta,t\rvert\omega)$, which denotes the probability density of the phases of oscillators with natural frequency $\omega$ at time $t$. The critical value $\lambda_c$ is the place where the trivial stationary distribution $\rho^*=1/(2\pi)$ (i.e., the oscillators spread uniformly on the unit circle) loses stability. For the fully-connected network, we first obtain the continuum limit of Eqs.~\eqref{eq:GeneralKuramoto}--\eqref{eq:MemoryDynamics}, see Eqs.~\eqref{MFE2}--\eqref{MFE5} in Methods.~We then derive and perform stability analysis of the trivial distribution to the delayed FPEs\cite{Frank2003}, Eq.~\eqref{DFPE1} in Methods, to approximate $\lambda_c$ for different cases.
\begin{figure*}[!t]
	\subfigure{\label{Fig4:a}}
	\subfigure{\label{Fig4:b}}
	\subfigure{\label{Fig4:c}}
	\subfigure{\label{Fig4:d}}
	\subfigure{\label{Fig4:e}}
	\subfigure{\label{Fig4:f}}
	\subfigure{\label{Fig4:g}}
	\subfigure{\label{Fig4:h}}
	\includegraphics[width=18cm]{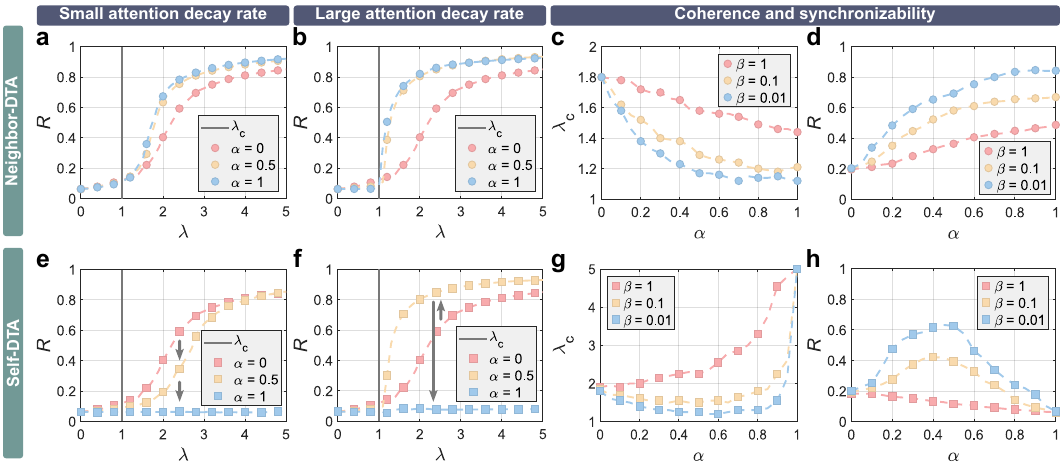}
	\caption{\label{Fig4} {\bf The effect of DTA in the WS networks.} Panels {\bf a}--{\bf d} and {\bf e}--{\bf h} show the results after applying the {\it neighbor-} and {\it self-DTA}. In all cases, the natural frequency distribution is set as $g(\omega)=\delta(0)$. {\bf a}, Order parameter $R$ for different coupling strength $\lambda$ under different attention weight (red: $\alpha=0$; yellow: $\alpha=0.5$; blue: $\alpha=1$) when attention decay rate $\beta=1$. {\bf b}, Order parameter $R$ versus $\lambda$ with the same settings as those in panel {\bf a} except that $\beta=0.01$. {\bf c}, Critical coupling strength $\lambda_c$ for different $\alpha$ under distinct $\beta$ (red: 1; yellow: 0.1; blue: 0.01). {\bf d}, Order parameter $R$ for different $\alpha$ with fixed $\lambda=1.5$. {\bf e}, Order parameter $R$ for different coupling strength $\lambda$ when $\beta=1$. {\bf f}, Order parameter $R$ versus $\lambda$ when $\beta=0.01$. {\bf g}, Critical coupling strength $\lambda_c$ for different $\alpha$. Non-monotonicity appears for small $\beta$ (yellow and blue). {\bf h}, Order parameter $R$ for different $\alpha$ with fixed $\lambda=1.5$. The non-monotonicity is consistent with that in panel {\bf g}.}
\end{figure*}

Before demonstrating the results, we emphasize that our {\em Emergence Transformer} framework advances the study of coupled phase oscillators by addressing two challenges.~First, we go beyond the analysis of distributed time delays from two-oscillator systems\cite{Ross2021} to large, complex networks as a framework incorporating DTA.~Second, when considering DTA as a type of memory, we drop the conventional, mathematically-convenient assumption of an infinite-memory integral on $[0, \infty]$.~Instead, we introduce a time-varying delay (or attention) kernel over $[0, t]$ (Methods).~This formulation is more physically plausible but poses significant analytical challenges.~Moreover, for the {\em self-DTA}, the varying kernel and inconsistent topologies ($\hat{\bm A}\neq{\bm A}$) yield highly intricate FPEs. Despite these challenges, we successfully derive accurate analytical approximations for the critical coupling strength $\lambda_c$.

For the {\it neighbor-DTA}, our result shows that $\lambda_c$ satisfies the following equation
\begin{equation}
	\lambda_c\int_{-\infty}^{+\infty} \frac{D}{D^2+\omega^2} g(\omega) {\rm d}\omega=2.\label{reslut1}
\end{equation}
Specifically, we have $\lambda_c=2D$ when $g(\omega)=\delta(0)$.~The result agrees well with the numerical simulation and is independent of attention weight $\alpha$ and decay rate $\beta$ (Figs.~\ref{Fig3:a}--\ref{Fig3:b}) suggesting that adding the {\it neighbor-DTA} does not affect synchronizability of fully-connected networks.~Moreover, we verify again that the coherence is also unchanged (Fig.~\ref{Fig3:a}).~As for the the {\it self-DTA}, $\lambda_c$ depend on both $\alpha$ and $\beta$ indicating the essential role of the {\it self-DTA}.~Specifically, we derive that $\lambda_c$ satisfies $\int_{-\infty}^{+\infty} {\rm Re}[f(0,\omega,\alpha,\beta,\lambda_c)] g(\omega) {\rm d}\omega = 2$, where the expression of $f(0,\omega,\alpha,\beta,\lambda)$ is given in Methods.~The analytical approximations also agree well with the numerical results and we observe that the {\it self-DTA} suppresses the synchronizability (Figs.~\ref{Fig3:c}--\ref{Fig3:d}).~When $g(\omega)=\delta(0)$, we further obtain explicitly $\lambda_c=2D[1-\alpha\beta/(\beta+4D)]^{-1}$. Apparently, larger attention weight $\alpha$ yield greater $\lambda_c$ (less synchronizability).~Moreover, the suppression of coherence is also verified again for different $\lambda>\lambda_c$ (Fig.~\ref{Fig3:c}).\\
\newline
\noindent{\bf Non-monotonic emergence in WS networks}\\

We next investigate networks for which mean-field theory is expected to fail.~By setting a low average degree and rewiring probability, we create sparse and structured WS networks (Methods). The dynamics of each oscillator are thus mainly influenced by its local feature rather than the global average.~This heterogeneity makes the mean-field approximation inadequate for predicting the synchronizability ($\lambda_c$). Consequently, we employ numerical simulations for this complicated case. Through these explorations, we systematically uncover the significant role of DTA by focusing on how attention weight $\alpha$ and its decay rate $\beta$ reshape the emergence behavior.

The results of the {\it neighbor-DTA} are shown in Figs.~\ref{Fig4:a}--\ref{Fig4:d}.~We find that, for both small and large decay rate ($\beta=0.01$ and $1$, respectively), applying the {\it neighbor-DTA} enhances both synchronizability (less critical coupling strength $\lambda_c$) and coherence (greater order parameter $R$), see Figs.~\ref{Fig4:a}--\ref{Fig4:b}.~We also compute $\lambda_c$ and $R$ for various attention weight $\alpha$ and decay rate $\beta$.~As $\alpha$ increases, $\lambda_c$ keeps decreasing (Fig.~\ref{Fig4:c}) and $R$ keeps increasing (Fig.~\ref{Fig4:d}). The two properties of synchronization are consistent for small variance of natural frequencies (Extended~Data~Fig.~\ref{ExFig1}) suggesting that the {\it neighbor-DTA} consistently promotes synchronization of the WS networks (with large ASPL).

For the {\it self-DTA}, the results are shown in Figs.~\ref{Fig4:e}--\ref{Fig4:h}. The qualitative behavior of the two properties now depends on the value of attention decay rate $\beta$.~When $\beta=1$ (relatively large), both properties vary monotonically (Fig.~\ref{Fig4:e}).~Larger attention weight $\alpha$ yields greater critical coupling strength $\lambda_c$ and less coherence $R$. More results are given in Figs.~\ref{Fig4:g}--\ref{Fig4:h} (red curves) for different values of $\alpha$.~As $\beta$ decreases and is sufficiently small, the monotonic variations become non-monotonic (Fig.~\ref{Fig4:b}).~More results on non-monotonic synchronizability and coherence are provided respectively in Figs.~\ref{Fig4:g} and \ref{Fig4:h} (blue and yellow curves).~Those results are consistent with those presented in Fig.~\ref{Fig2}.~More importantly, when applying the {\it self-DTA}, the variations in synchronizability and coherence are always consistent.~The above findings establish fundamental principle of modulating the emergence phenomenon of synchronization flexibly through attention.\\
\newline
\noindent{\bf Emergence in real-world networks}\\

Based on the results obtained so far, we know that applying the {\it neighbor-DTA} consistently promotes the emergence of synchronization and coherence.~Contrarily, the {\it self-DTA} plays a more complicated role depending on the spatial network topology.~This is because networks have distinct efficiency in transmitting spatial information, which is highly related to ASPL. Particularly, this value of the fully-connected network is 1, while it is approximately $6$ for the WS networks considered in previous examples.~We thus hypothesize that the non-monotonic variation occurs when the network has less transmission efficiency (higher ASPL). 
\begin{figure}[!t]
	\subfigure{\label{Fig5:a}}
	\subfigure{\label{Fig5:b}}
	\subfigure{\label{Fig5:c}}
	\subfigure{\label{Fig5:d}}
	\includegraphics[width=8.8cm]{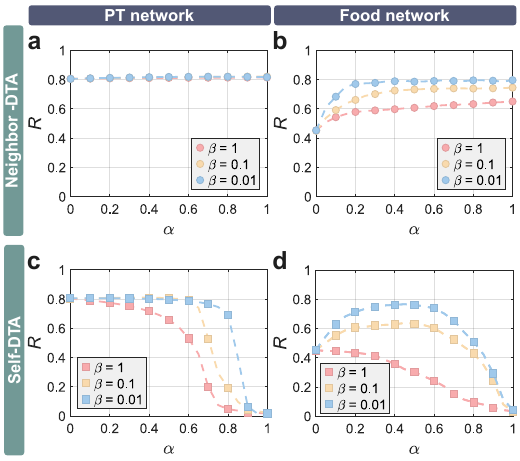}
	\caption{\label{Fig5} {\bf The effect of DTA on opinion dynamics with real-world networks.} All the panels show the results of order parameter $R$ for different attention weight $\alpha$ with fixed coupling strength $\lambda=1.5$ and distinct attention decay rate $\beta$ (red: 1; yellow: 0.1; blue: 0.01). Panels {\bf a} and {\bf b} correspond respectively to the PT and food networks with the {\it neighbor-DTA}. Panels {\bf c} and {\bf d} correspond respectively to the PT and food networks with the {\it self-DTA}. }
\end{figure}

To test our hypothesis when applying the {\it self-DTA}, we conduct more experiments with diverse network topologies, whose ASPL are provided in Extended~Data~Tab.~\ref{ExTab1}.~We first consider two representative and common networks: the Erdős--Rényi (ER) network \cite{Erdos1960} and the Barabási--Albert (BA) network \cite{Barabasi1999}, see Extended~Data~Fig.~\ref{ExFig2}.~The ASPL of the ER network is approximately~1.5 that is close to the fully-connected one. Thus, synchronizability is monotonically suppressed as more and more {\it self-DTA} are applied. As for the BA network whose ASPL is approximately 3.3, the results look similar to those of the WS network.~When the attention decay rate $\beta$ is small enough (e.g., $\beta=0.01$), non-monotonic variation occurs.~Particularly, applying adequate amount of {\it self-DTA} optimally promotes the emergence of synchronization. 
\begin{figure*}[!t]
	\subfigure{\label{Fig6:a}}
	\subfigure{\label{Fig6:b}}
	\subfigure{\label{Fig6:c}}
	\subfigure{\label{Fig6:d}}
	\subfigure{\label{Fig6:e}}
	\subfigure{\label{Fig6:f}}
	\subfigure{\label{Fig6:g}}
	\subfigure{\label{Fig6:h}}
	\includegraphics{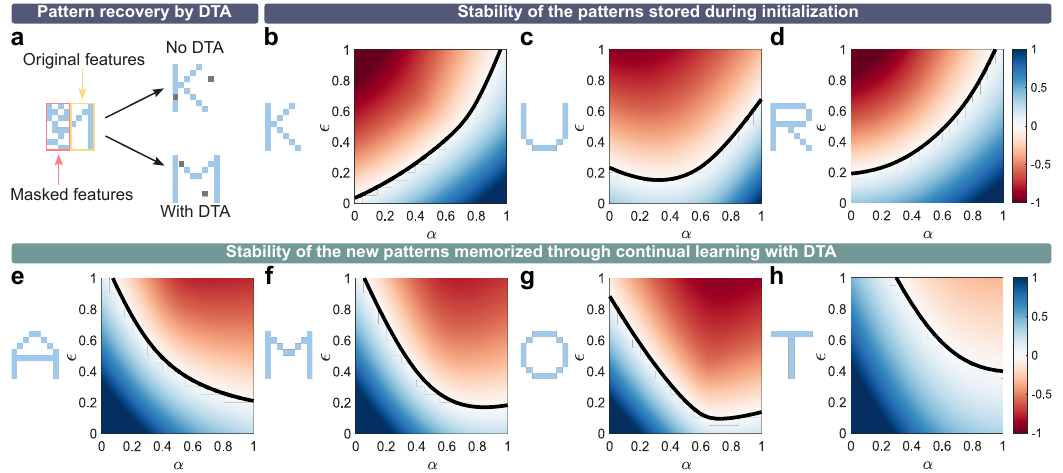}
	\caption{\label{Fig6} {\bf The effect of DTA for the emergence of continual learning in the HNN.} {\bf a}, The HNN is designed for recovering the masked pattern (red: masked; yellow: original). The letter is not recovered without attention. By applying appropriate DTA, the recovery is successful. We consider 7 letters `KURAMOT' in this work. The rest of panels show, in $\epsilon$-$\alpha$ parameter space, the leading (right-most) eigenvalues of the Jacobian matrix of the synchronization state corresponding to each letter: {\bf b}, `K'; {\bf c}, `U'; {\bf d}, `R'; {\bf e}, `A'; {\bf f}, `M'; {\bf g}, `O'; {\bf h}, `T'. The heatmap show the stable region (red) and the unstable region (blue) for each pattern. The black solid curve represent the boundary between stable and unstable regions.}
\end{figure*}

We also analyze the role of DTA in real-world networks by considering the emergent coherence of opinions in a variant of the Kuramoto phase model proposed recently\cite{Ojer2023}.~See Methods for model details.~In these models, synchronization represents the consensus of opinions.~Moreover, applying DTA means that each individual focuses on the observed information that is a practically-natural mechanism.~We explore two real social networks called Portuguese Twitch (PT) and Food networks\cite{Rossi2015}.~For both networks, utilizing the {\it neighbor-DTA} promotes synchronization (Figs.~\ref{Fig5:a}--\ref{Fig5:b}). As for the {\it self-DTA}, the results still depend on ASPL, being consistent with our findings.~The ASPL of the PT and Food networks are close to those of the fully-connected and the WS networks, respectively (Extended~Data~Tab.~\ref{ExTab1}).~Consequently, the {\it self-DTA} suppresses the coherence of the PT network, whereas there is an optimal attention weight for promoting the opinion consensus in the Food network (Figs.~\ref{Fig5:c}--\ref{Fig5:d}).
\\
\newline
\noindent{\bf Achieving emergent continual learning}\\

We finally demonstrate an application of DTA to a paradigmatic machine learning model HNN, also known as an associative memory network.~It is an oscillatory neural network for recovering masked patterns by memorize them as different attractors\cite{Fu2025}.~The information are stored in a customized network topology ${\bm C}=\{C_{ij}\}$.~When using the Kuramoto-like model, those attractors are considered to be different coherent states, see Methods for more details.~When the (spatial) network topology is fixed, we can only make the HNN memorize or forget the old patterns by changing the coupling strength (e.g., `KUR' in Fig.~\ref{Fig6}).~In light of our findings, the ability of memorizing new patterns emerges by leveraging DTA through an attention network $\hat{\bm C}=\{\hat{C}_{ij}\}$ which is added with certain attention weight $\alpha$ (Fig.~\ref{Fig6:a}). As the two topologies are different, the attention network $\hat{\bm C}$ suppresses the information of the old pattern stored in the spatial network ${\bm C}$ (Figs.~\ref{Fig6:b}--\ref{Fig6:d}).~Consequently, as $\alpha$ increases, the HNN gradually forgets the symbols `KUR'. Contrarily, it memorizes the new patterns `AMOT' (Figs.~\ref{Fig6:e}--\ref{Fig6:h}). Moreover, the leading eigenvalue of those new attractors exhibit non-monotonic behavior as $\alpha$ increases because of the difference between $\hat{\bm C}$ and ${\bm C}$. Those results are consistent with those findings of the {\it self-DTA} where the attention and spatial networks are different.~We acknowledge that the underlying mechanism of the HNN with DTA might be more complicated, but we still provide a flexible way to modulate its pattern-recovering capability. \\
\newline
\noindent{\bf Discussion and prospects}\\

The attention mechanism in the Transformer architecture, leveraging non-local information of sequential data, has led to great achievements of machine learning\cite{Vaswani2017,Lin2022}. Inspired by this non-local feature, here, we develop the {\em Emergence Transformer} for exploring the role of DTA, a time-varying attention on temporal sequence, in flexibly modulating the emergence phenomena, especially, the coherence or incoherence of components in complex systems.~We incorporate DTA into the concerned dynamical model by carefully designing dynamical query, key and value matrices, i.e., ${\bm Q}$, ${\bm K}$ and ${\bm V}$. The essential difference between our matrices with those in the classical Transformer is their dynamical property. Specifically, the three matrices change persistently as time evolves. As such, we surprisingly find that those classical attention in LLMs is a particular case of the DTA in our framework.~While the parameter matrices can be customized or trained, we focus on two natural scenarios.~They are {\em neighbor-} and {\em self-DTA}, where each component utilizes the non-local temporal sequence of its neighbors or its own, respectively.

We also uncover the dual roles of the DTA in the context of synchronization.~Our main finding reveals that the {\it neighbor-DTA} consistently promotes (or at least keeps) the emergence of synchronization.~Moreover, the effect of the {\it self-DTA} depends tightly on the information-transmission efficiency, as quantified by ASPL of networks. Specifically, it acts as a universal suppressor when APLE is low. As for higher ASPL, the {\it self-DTA} exhibits a non-monotonic behavior, enabling an optimal attention weight to enhance the emergence of synchronization. Our findings establish fundamental principles for designing a flexible and feasible modulation strategy. Instead of altering the original network topology that is often expensive or inaccessible, we simply achieve it by storing and manipulating past states of temporal sequence. The applicability of those principles extends to fields like artificial intelligence, suggesting a mechanism for enhancing continual learning. By using DTA appropriately, it allows new patterns to be learned without catastrophic forgetting.

Besides immediate practical implications, we also overcome theoretical challenges in analyzing nonlinear dynamics of complex systems.~Actually, the proposed {\em Emergence Transformer} is inherently characterized by a set of DDEs with distributed delays\cite{Frank2005,Lee2009,Zhang2021}.~Those attention matrices are updated dynamically, therefore, the attention kernel is time varying and is thus more physically plausible than those infinite kernels.~Our attention kernel does not require to know infinite past states.~Additionally, we bridge the gap between the studies on two components and the dynamics\cite{Ross2021} of large, complex networks with the presence of distributed time delays.~One more contribution of the present study is the development of the analytical framework for approximating the critical coupling strength from the Fokker-Planck equation with distributed and time varying delays.~This framework has the potential for exploring the role of DTA in a much broader class of emergence phenomena in complex systems, including synchronization with significant amplitude dynamics\cite{Zhang2021}.

Our explorations of the phase dynamics demonstrate the efficacy of the {\em Emergence Transformer}.~In fact, the framework admits generalization to more complex settings through trainable parameter matrices, suggesting its applicability to a wider range of emergence phenomena.~For example, we may consider the emergence of synchronization in phase-amplitude dynamics. Accordingly, we can extend the parameter matrices to complex field as ${\bm W}^{\bm Q}\in\mathbb{C}^{N\times d}$, ${\bm W}^{\bm K}\in\mathbb{C}^{N\times d}$ and ${\bm W}^{\bm V}\in\mathbb{C}^{T\times T}$. Actually, the {\em Emergence Transformer} characterized by Eq.~\eqref{eq:GeneralKuramoto} can also be derived from the phase-amplitude equations under the same conditions as those in the Kuramoto's work\cite{Kuramoto1975} (see Methods and SI Secs.~8--9). 

Our findings also open new avenues for future studies because DTA reminds us natural memory mechanism and spatiotemporal information exchange.~In synthetic biology, for instance, our framework could guide the design of RNA-based circuits that encode memory for controlling emergent cellular behaviors\cite{Cai2025}.~It also benefits the development of synchronization-based quantum communication systems\cite{Schmolke2022} with non-Markovian effects\cite{Vega2017}.~From a theoretical viewpoint, the discovered interplay between spatial and temporal information suggests a connection to the Takens embedding theorem\cite{Takens1981,Ma2018,Raut2025}.~Exploring the underlying relation may provide a rigorous foundation for designing highly efficient attention protocols from just a single node's time series but being able to control network-wide emergence phenomenon. Finally, the framework developed here is not confined to emergent coherence and is extendable to explore the role of DTA in other emergence phenomenon and collective behaviors in complex systems\cite{LiuJ2024}.


\newpage
\clearpage
\section{Methods}
\noindent{\bf DTA in a discrete-time model}\\

\noindent In order to have a better understanding and comparison with the classical Transformer architecture, we also demonstrate how to derive attention information for updating the phase states when time is discrete, see also Fig.~\ref{Fig1:a}. The general discrete-time model for $N$ phase oscillator reads $\theta^{(i)}_{T+1}=f_i({\bm \Theta}_T,{\bm M}_T),~i=1,2,\dots,N$. The functions $f_i$ is the updating rule and we do not focus on them in this work. We are interested in the phase matrix ${\bm \Theta}_T\in\mathbb{R}^{T\times N}$ and the attention matrix (vector) ${\bm M}_T\in\mathbb{R}^{1\times N}$.

The element in the $i$th row and the $j$th column of ${\bm \Theta}_T$ is ${\rm e}^{\mi\theta^{(j)}_i}$ representing the complex state of the $j$th oscillator at time $i$.~Analogous to the continuous-time case, the size of ${\bm \Theta}_T$ changes after each iteration (e.g., $T$ changes to $T+1$ after one iteration).~To compute ${\bm M}_T$, we also pre-design or learn appropriate parameter matrices ${\bm W}^{\bm Q}\in\mathbb{R}^{N\times d}$, ${\bm W}^{\bm K}\in\mathbb{R}^{N\times d}$ and set ${\bm W}^{\bm V}={\bm I}_N\in\mathbb{R}^{N\times N}$, see Fig.~\ref{Fig1:a}. Then, we obtain the attention kernel as the last row of the matrix ${\bm C}={\rm Softmax}\left(\frac{|{\bm Q}\bar{\bm K}^\top|}{\sqrt{d}}\right)\in\mathbb{R}^{T\times T}$. Finally, the attention vector used to update the phase states is computed as ${\bm M}_T=\sum_{k=1}^T C_{Tk}{\bm \theta}_k$, where ${\bm \theta}_k$ is the $k$th row of ${\bm \Theta}_T$. Actually, we have ${\bm M}_T=\left[M_T^{(1)},M_T^{(2)},\dots,M_T^{(N)}\right]$, where $M_T^{(i)}$ is the attention information of the $i$th oscillator. Once the attention vector ${\bm M}_T$ is obtained, we update the phase states for the next timestep, like predicting the next token in LLMs. See SI~Sec.~3 for more details.\\

\noindent{\bf Mean-field equations}\\

\noindent To obtain the critical value $\lambda_c$, we consider the mean-field equations when $N\rightarrow\infty$ for the fully-connected network where $A_{ij}=1-\delta_{ij}$. In the continuum limit, the states of oscillators are described by the probability density function $\rho(\theta,t|\omega)$. It satisfies the normalization condition $\int_{0}^{2\pi} \rho(\varphi,t|\omega){\rm d}\varphi=1$. Then, the order parameter $r_t$ for the mean-field equation is defined as
\begin{equation}\nonumber
	r_t(\rho){\rm e}^{{\rm{i}}\psi_t(\rho)}\coloneqq\int^{+\infty}_{-\infty}\int_0^{2\pi}{\rm e}^{{\rm{i}}\varphi} \rho(\varphi,t | \omega)g(\omega) {\rm d}\varphi {\rm d}\omega. 
\end{equation}

Now, the mean-field equation for the oscillator with natural frequency $\omega$ at the continuum limit is written as
\begin{equation}
	\dot{\theta}_t= \omega+\lambda I(\theta_t,M_t)+\xi_t, \label{MFE2}
\end{equation}
where $I(\theta_t,M_t)$ represents the total information received at time $t$. It admits the following form
\begin{equation}
	I(\theta_t,M_t)=(1-\alpha)r_t\sin(\psi_t-\theta_t)+\alpha{{\rm{Im}}}[M_t {\rm e}^{-{\rm{i}}\theta_t}]. \label{MFE3}
\end{equation}
The difference between the {\it neighbor-DTA} and the {\it self-DTA} appear in the dynamics of $M_t$.~The information of {\it neighbor-DTA} evolves as
\begin{equation}
	\dot{M}_{t,{\rm n}}= \beta\left(r_t{\rm e}^{{\rm{i}}\psi_t}-M_{t,{\rm n}}\right), \label{MFE4}
\end{equation}
while the {\it self-DTA} information evolves as
\begin{equation}
	\dot{M}_{t,{\rm s}}=  \beta\left({\rm e}^{{\rm{i}}\theta_t}-M_{t,{\rm s}}\right). \label{MFE5}
\end{equation}
In Eqs.~\eqref{MFE4} and \eqref{MFE5}, the subscripts ${\rm n}$ and ${\rm s}$ represent neighbor and self, respectively.\\

\noindent{\bf Delayed FPE}\\

\noindent From the models at the continuum limit, we derive the delayed FPEs for both {\it neighbor-} and {\it self-DTA}. Here, we provide main procedures and the detailed derivations are given in SI~Secs.~5--6. Both equations read
\begin{equation}
	\frac{\partial \rho}{\partial t}=D \frac{\partial^2 \rho}{\partial \theta^2}-\omega\frac{\partial \rho}{\partial \theta}-\lambda\frac{\partial}{\partial \theta}[I(\theta,M_{t,a})\rho],\label{DFPE1}
\end{equation}
where $a\in\{{\rm n}, {\rm s}\}$. Solving $M_{t,a}$ from Eqs.~\eqref{MFE4} and \eqref{MFE5}, we obtain
\begin{equation}\nonumber
	\begin{aligned}
		I(\theta,M_{t,a})=&~(1-\alpha)r_t(\rho)\sin[\psi_t(\rho)-\theta]\\
		&+\alpha\, \hat{r}_{t,a}(\rho)\sin[\hat{\psi}_{t,a}(\rho)-\theta],
	\end{aligned}
\end{equation}
where, for the {\it neighbor-DTA},
\begin{equation}\nonumber
	\begin{aligned}
		\hat{r}_{t,{\rm n}}(\rho){\rm e}^{{\rm{i}}\hat{\psi}_{t,{\rm n}}(\rho)}=& ~ {\rm e}^{-\beta t}r_0(\rho){\rm e}^{{\rm i}\psi_0(\rho)}\\
		&+\beta{\rm e}^{-\beta t}\int^t_0{\rm e}^{\beta s}r_s(\rho){\rm e}^{{\rm i}\psi_s(\rho)} {\rm d}s,
	\end{aligned}
\end{equation}
and for the {\it self-DTA}
\begin{equation}\nonumber
	\begin{aligned}
		\hat{r}_{t,{\rm s}}(\rho){\rm e}^{{\rm{i}}\hat{\psi}_{t,{\rm s}}(\rho)}=& ~ {\rm e}^{-\beta t}\int_0^{2\pi}{\rm e}^{{\rm{i}}\varphi} \rho(\varphi,s|\theta ,t,\omega){\rm d}\varphi \\
		&+\int^t_0\int_0^{2\pi}\beta {\rm e}^{\beta(s-t)}{\rm e}^{{\rm{i}}\varphi} \rho(\varphi,s|\theta ,t,\omega){\rm d}\varphi {\rm d}s,
	\end{aligned}
\end{equation}
where the conditional distribution $\rho(\varphi,s|\theta ,t,\omega)$ with~$s\leqslant t$~denotes the probability that an oscillator with natural frequency $\omega$ has phase $\varphi$ at time $s$ given that the phase of the oscillator at time $t$ is $\theta$. The joint distribution is given by $\rho(\varphi,s;\theta ,t|\omega) = \rho(\varphi,s|\theta ,t,\omega) \times \rho(\theta,t|\omega)$.\\ 
\newline
\noindent{\bf Stability analysis:}~{\it\textbf{neighbor-DTA}}\\

\noindent The synchronization occurs when the trivial distribution $\rho^*$ of Eq.~\eqref{DFPE1} loses stability. Therefore, we perform the stability analysis by perturbing $\rho^*$ as $\rho=\rho^*+\varepsilon\eta$ with $0<\varepsilon\ll1$ and analyzing the stability of zero solution to the equation of $\eta$. We have $\rho^*(\theta|\omega)=1/(2\pi)$, and therefore $	\rho(\theta,t|\omega)=1/(2\pi)+\varepsilon \eta(\theta,t|\omega)$ with  $\int^{2\pi}_{0}\eta(\varphi,t|\omega){\rm d}\varphi=0$. From Eq.~\eqref{DFPE1}, we then obtain the equation at leading order $\mathcal{O}(\varepsilon)$ for perturbation $\eta(\theta,t|\omega)$ as 
\begin{equation}\nonumber
	\begin{aligned}
		\frac{\partial \eta}{\partial t}=&~D \frac{\partial^2 \eta}{\partial \theta^2}-\omega\frac{\partial \eta}{\partial \theta}+\frac{\lambda}{2\pi}(1-\alpha) r_{t}(\eta) \cos[\psi_{t}(\eta)-\theta] \\
		&+\frac{\lambda}{2\pi}\alpha\,\hat{r}_{t,{\rm n}}(\eta) \cos[\hat{\psi}_{t,{\rm n}}(\eta)-\theta].
	\end{aligned}
\end{equation}
We then apply the Fourier decomposition as $\eta(\theta,t|\omega)=\frac{1}{2\pi}\sum^\infty_{\ell=0}\left[c_\ell(t|\omega){\rm e}^{{\rm i}\ell\theta}+{\rm c.c.}\right]$ where c.c.~stands for the complex conjugate.~Considering the order parameter, we note that $r_t(\eta)=|c_1(t)|$. Therefore, we then obtain the equations of $c_1(t|\omega)$ and analyze the stability of the zero solution.~We finally deduce that $\lambda_c$ satisfy the following equation
$$\lambda_c\int^{+\infty}_{-\infty}\frac{D}{D^2+\omega^2}g(\omega){\rm d}\omega=2.$$ 
See SI~Sec.~5.2 for more detailed derivations.\\

\noindent{\bf Stability analysis:}~{\it\textbf{self-DTA}}\\

\noindent For the {\it self-DTA}, we have $\rho^*(\varphi;\theta)=1/(4\pi^2)$, and therefore, we perform the perturbation as $\rho(\varphi,s;\theta,t|\omega)=1/(4\pi^2)+\varepsilon\zeta(\varphi,s;\theta,t|\omega)$. Note that we have $\int^{2\pi}_{0}\zeta(\varphi,s;\theta,t|\omega) {\rm d}\varphi=\eta(\theta,t|\omega)$ and $\int^{2\pi}_{0}\zeta(\varphi,s;\theta,t|\omega) {\rm d}\theta  =\eta(\varphi,s|\omega)$. From Eq.~\eqref{DFPE1}, we then obtain the equations at order $\mathcal{O}(\varepsilon)$ for $\eta(\theta,t|\omega)$ as
\begin{equation}
	\begin{aligned}
		\frac{\partial \eta}{\partial t}&=~D \frac{\partial^2 \eta}{\partial \theta^2}-\omega\frac{\partial \eta}{\partial \theta}+\frac{\lambda}{2\pi}(1-\alpha) r_t(\eta) \cos[\psi_t(\eta)-\theta] \\
		&+\alpha\lambda{\rm e}^{-\beta t}\bigg\{Y(0)+Z(0)+\int^t_0\beta{\rm e}^{\beta s}[Y(s)+Z(s)]{\rm d}s\bigg\},\label{SA3}  
	\end{aligned}
\end{equation}
where
\begin{equation}\nonumber
		\begin{aligned}
			Y(s) = & ~ \int^{2\pi}_{0} \cos(\varphi-\theta)\zeta(\varphi,s;\theta,t|\omega){\rm d}\varphi,\\
			Z(s) = & ~\int^{2\pi}_{0} -\sin(\varphi-\theta)\frac{\partial}{\partial \theta}\zeta(\varphi,s;\theta,t|\omega){\rm d}\varphi.
		\end{aligned}
\end{equation}

Then, we also apply the Fourier decomposition. The decomposition of $\eta(\theta,t|\omega)$ is the same as that of the {\it neighbor-DTA}.~The decomposition of $\zeta(\varphi,s;\theta,t|\omega)$ is written as
\begin{equation}\nonumber
		\begin{aligned}
			\zeta(\varphi,s;\theta,t|\omega)=& ~ \frac{1}{4\pi^2}\sum^{\infty}_{k=-\infty}\sum^{\infty}_{\ell=-\infty}c_{k,\ell}(s,t|\omega){\rm e}^{{\rm{i}}\ell\theta}{\rm e}^{{\rm{i}}k\varphi}.
		\end{aligned}
\end{equation}
Analogously, we need to analyze the dynamics of $c_1(t|\omega)$ whose equation is obtained from Eq.~\eqref{SA3} as
\begin{equation}\nonumber
	\begin{aligned}
		\frac{\partial c_1}{\partial t}= & ~-(D+{\rm{i}}\omega) c_1+\frac{(1-\alpha)\lambda}{2}\int_{-\infty}^{+\infty}c_1(t|\omega)g(\omega){\rm d}\omega\\
		& ~ +\frac{\alpha\lambda}{2 }{\rm e}^{-\beta t}\left[c_1(0|\omega)-c_{-1,2}(0,t|\omega)\right]\\
		& ~ +\frac{\alpha\lambda}{2 }\int^t_0 \beta {\rm e}^{\beta(s-t)}\left[c_1(s|\omega)-c_{-1,2}(s,t|\omega)\right]{\rm d}s,
	\end{aligned}
\end{equation}
which is a DDE. 

In order to analyze the evolution of $c_1(t|\omega)$ we need to know $c_{-1,2}(s,t|\omega)$. To depict their relationship, we derive the FPE of $\rho(\varphi,s;\theta,t|\omega)$ and the perturbation equation of $\zeta(\varphi,s;\theta,t|\omega)$ (see SI~Sec.~6 for those equations). To simplify the computation and analysis, we only consider the leading term of the equation for $\zeta(\varphi,s;\theta,t|\omega)$, which is $\frac{\partial \zeta}{\partial t}=D \frac{\partial^2 \zeta}{\partial \theta^2}-\omega\frac{\partial \zeta}{\partial\theta}$.
The accuracy of this approximation are verified through numerical simulations (Fig.~S2). Applying the Fourier decomposition and considering the coefficients of ${\rm e}^{-{\rm{i}}\varphi}{\rm e}^{2{\rm{i}}\theta}$, we obtain the equation of $c_{-1,2}$ as
\begin{equation}\nonumber
	\frac{\partial }{\partial t}c_{-1,2}(s,t|\omega)=-4D c_{-1,2}(s,t|\omega)-2{\rm i}\omega c_{-1,2}(s,t|\omega).
\end{equation}
Note that the initial condition is $c_{-1,2}(s,s|\omega)=c_{1}(s|\omega)$, and we thus obtain 
$$	c_{-1,2}(s,t|\omega)={\rm e}^{-(4D+2{\rm i}\omega)(t-s)}c_1(s|\omega),$$
which allows us to obtain an equation of $c_1(t|\omega)$ as
\begin{equation}\nonumber
	\begin{aligned}
		\frac{\partial c_1}{\partial t}=& -(D+{\rm{i}}\omega) c_1+\frac{(1-\alpha)\lambda}{2}\int_{-\infty}^{+\infty}c_1(t|\omega)g(\omega){\rm d}\omega\\
		&+\frac{\alpha\lambda}{2}{\rm e}^{-\beta t}\left[1-{\rm e}^{-(4D+2{\rm i}\omega)t}\right]c_1(0|\omega)\\
		&+\frac{\alpha\lambda}{2}\int^t_0 \beta {\rm e}^{\beta(s-t)}\left[1-{\rm e}^{(4D+2{\rm i}\omega)(s-t)}\right]c_1(s|\omega){\rm d}s. 
	\end{aligned}
\end{equation}

Finally, by analyzing the stability of the zero solution, we deduce that the critical coupling strength $\lambda_c$ is approximated by the real solution of 
\begin{equation}\nonumber
	\int_{-\infty}^{+\infty} {\rm Re}[f(0,\omega,\alpha,\beta,\lambda_c)] g(\omega) {\rm d}\omega = 2,
\end{equation}
where $$f(0,\omega,\alpha,\beta,\lambda) \coloneqq \dfrac{(1-\alpha)\lambda}{D+{\rm{i}}\omega -\dfrac{\alpha\lambda}{2}\dfrac{(4D+2{\rm i}\omega)}{(\beta+4D+2{\rm i}\omega)}}.$$ When $\alpha=1$, there is no synchronization for any coupling strength $\lambda$ since $1-\alpha=0$. We provide more detailed derivations in SI~Sec.~6.2.\\

\noindent{\bf Opinion dynamics}\\

\noindent For real-world applications, we consider the opinion dynamics in real-world networks. The model is a variant of the Kuramoto model that is written as 
\begin{equation}\nonumber
	\dot{\theta}_t^{(i)}=\rho^{(i)}\sin[\varphi^{(i)}-\theta^{(i)}_t]+\lambda{\rm Im}[\mathcal{I}^{(i)}_t{\rm e}^{-{\rm i}\theta^{(i)}_t}]+\xi^{(i)}_t.
\end{equation}
It is inspired by the Friedkin--Johnsen model\cite{Friedkin1990,Ojer2023}. For each individual $i$, the phase $\theta_t^{(i)}$ characterizes the orientation of its opinion at time $t$. The term $\varphi^{(i)}$ represents its inherent opinion. When there is no interactions, each individual's opinion converges to its inherent one at a certain speed controlled by $\rho^{(i)}$. The coupling term and noise are the same as those in Eq.~\eqref{eq:GeneralKuramoto}.~When $\rho^{(i)} \equiv 0$, the model is reduced to Eq.~\eqref{eq:GeneralKuramoto} with $g(\omega)=\delta(0)$. In a real-world scenario, social influence manifest as the tendency of individuals to update their opinion (through the interactions) to meet the expectations of their neighbors, which make the opinions achieve a consensus (or coherent) state when the coupling strength $\lambda$ is large. Such a consensus state is actually the emergent synchronous state that we are interested in. \\

\noindent{\bf Continual learning of the HNN}\\

\noindent The HNN leverages the associative memory to store patterns. For the classical HNN using the phase oscillator, each binary pattern is encoded as an attractor (i.e., equilibrium) of the dynamical system \cite{Fu2025}. To illustrate how DTA affects the stability of those attractors, we consider a model incorporating attention mechanism that consists of $N$ identical coupled oscillators. The model reads
\begin{equation}\nonumber
	\dot{\theta}^{(i)}_t= {\rm Im}[\mathcal{I}_t^{(i)}{\rm e}^{-{\rm i}\theta_t^{(i)}}]+\epsilon{\rm Im}[\hat{R_t}{\rm e}^{-2{\rm i}\theta_t^{(i)}}],~i=1,2,\dots,N.
\end{equation}
There is no natural frequency for each oscillator, and there are two types of couplings. Specifically, pairwise and second-order interactions are characterized by the first and second terms on the right-hand side, respectively. Moreover, $\hat{R}_t=|N^{-1}\sum_{j=1}^N{\rm e}^{2{\rm i}\theta_t^{(j)}}|$ is the second-order parameter and $\epsilon$ is the strength of second-order coupling term.

The fist-order interaction term $\mathcal{I}^{(i)}_t$ includes the spatial information and attention information, which has the following form
\begin{equation}\nonumber
	\mathcal{I}_t^{(i)}=(1-\alpha)\sum_{j=1}^{N}C_{ij}{\rm e}^{{\rm i}\theta_t^{(j)}}+\alpha\sum_{j=1}^{N}\hat{C}_{ij}M_t^{(j)},
\end{equation}
where $\alpha$ still denotes the attention weight. The dynamics of the attention $M_t^{(j)}$ follows Eq.~\eqref{eq:MemoryDynamics}. When $\alpha=0$, the model reduces to the classical associative-memory network\cite{Nishikawa2004}.

The system has $2^N$ fixed points, up to translating every oscillator by a constant phase, corresponding to phase-locked solutions $|\theta^{(j)}-\theta^{(i)}| = 0$ or $|\theta^{(j)}-\theta^{(i)}| = \pi$ for all $i$ and $j$. The attractors consist of all stable fixed points\cite{Cornelius2013}. Considering the fixed point $\theta^{(i)}=0$ or $\pi$ after a phase shift, the corresponding vector ${\bm \eta}= [\eta_1,\eta_2,\dots,\eta_N]^\top=[{\rm e}^{{\rm i}\theta^{(1)}},{\rm e}^{{\rm i}\theta^{(2)}},\dots,{\rm e}^{{\rm i}\theta^{(N)}}]^\top$ consists of $-1$ and 1. In this way, each fixed point encodes uniquely a binary pattern. In order to stabilize (i.e., memorize) the desired states, the network stores the patterns in the network topology by Hebb's learning rule $C_{ij} = \frac{1}{N}\sum^{p}_{\mu=1}\eta^{(\mu)}_i\eta^{(\mu)}_j$, where $\boldsymbol{\eta}^{(\mu)}=[\eta^{(\mu)}_1,\dots,\eta^{(\mu)}_N]^\top$ with $\eta^{(\mu)}_i=\pm 1 ~ (i=1,\dots,N,\mu=1,\dots,p)$ is the set of $p$ binary patterns of length $N$ to be stored through spatial information\cite{Hoppensteadt1997}. Here, we introduce a network for applying the DTA. Specifically, $\hat{C}_{ij} = \frac{1}{N}\sum^{p+q}_{\mu=p+1}\eta^{(\mu)}_i\eta^{(\mu)}_j$ with $\eta^{(\mu)}_i=\pm 1 ~ (i=1,\dots,N,\mu=p+1,\dots,p+q)$ is the set of $q$ more binary patterns of length $N$. Those $q$ patterns are the ones need to be memorized by the original network without making the network forget the old patterns. Therefore, it is a kind of continual learning task. 

To analyze the stability of the attractors of those patterns, we consider the eigenvalue of the Jacobian matrix. Note that we always have one eigenvalue $\lambda=0$ corresponding to the synchronous manifold. Except  this eigenvalue, we focus on the leading eigenvalue which is the right most one on the complex plane. The real part of the leading eigenvalue characterizes the stability of the pattern ${\bm \eta}$: negative and positive for stable and unstable, respectively. Indeed, the Jacobian matrix and the leading eigenvalue are analytically computable, see SI~Sec.~7. 

In this work, we consider patterns of size $N=64$ including $p=3$ old and $q=4$ new patterns. All the patterns represent seven letters. The old patterns are `K', `U' and `R', while the new patterns are `A', `M', `O' and `T'. We analyze the stability of those patterns after adding the new ones through attention mechanism in the $\epsilon$--$\alpha$ parameter space. \\
\newline
\noindent{\bf DTA of phase-amplitude oscillators}\\

\noindent In this work, we consider the effect of DTA in coupled phase oscillators. It is also possible to study it in the coupled phase--amplitude oscillators where each individual $z_t^{(i)}\in\mathbb{C}$ is a Stuart--Landau oscillator. The model reads
\begin{equation}\nonumber
	\dot {z}_t^{(i)}=[{\rm i}\omega^{(i)}+a]z_t^{(i)}-b|z_t^{(i)}|^2z_t^{(i)}+\lambda \mathcal{Q}_t^{(i)}+w_t^{(i)},
\end{equation}
where $\omega^{(i)}$ is the natural frequency of the $i$th oscillator, $a$ and $b$ are positive real constants, and $\lambda$ is the coupling strength. The term $\mathcal{Q}_t^{(i)}$ represents the coupling. Moreover, $w_t^i$ stands for independent two-dimensional Gaussian white noise that satisfies $\mathbb{E}[w_t^{(i)}]= 0,~ {\rm Cov}[w_t^{(i)}w_s^{(j)}]= 2\sigma\delta_{ij}\delta(t-s){\bm I}_2$. Here, $\sigma$ denotes the noise strength. The coupling term $\mathcal{Q}_t^{(i)}$ describes the information received by the $i$th oscillator at time $t$. It is the weighted average of the spatial and the dynamical attention information that is written as
\begin{equation}
	\mathcal{Q}_t^{(i)}=(1-\alpha)\frac{1}{d_i}\sum_{j=1}^{N}A_{ij}z_t^{(j)}+\alpha\frac{1}{\hat{d}_i}\sum_{j=1}^{N}\hat{A}_{ij}Q_t^{(j)}.\label{normalKM2}
\end{equation}
In Eq.~\eqref{normalKM2}, $z_t^{j}$ and $Q_t^{(j)}$ represents, respectively, the spatial and the dynamical attention information of the $j$th oscillator at time $t$. Also, $\alpha$ is the attention weight. Analogous to the coupled phase oscillator, by assuming continuous accumulation with an exponential decay rate $\beta$, the dynamics of $Q_t^{(i)}$ is written as 
\begin{equation}\nonumber
	\dot{Q}_t^{(i)}=\beta\left[z_t^{(i)}-{Q_t^{(i)}}\right].
\end{equation}
The physical meanings of other parameters, $A_{ij}$, $d_i$, $\hat{A}_{ij}$ and $\hat{A}_{ij}$, are the same as those in Eq.~\eqref{eq:TotalInformation}. Besides the accumulation and decay, the attention can also be derived by manipulating appropriate complex-valued parameter matrices ${\bm W}^{\bm Q}\in\mathbb{C}^{N\times d}$, ${\bm W}^{\bm K}\in\mathbb{C}^{N\times d}$ and ${\bm W}^{\bm V}\in\mathbb{C}^{N\times N}$.

Note that the phase model considered in the present work can be reduced from the phase-amplitude model by considering appropriate assumptions (i.e. $a,b\to\infty$ and $a/b$ admits a finite limit) as in the work of Kuramoto\cite{Kuramoto1975}. We provide more detailed derivation in SI~Sec.~8. As the two models incorporate noise, we also consider the reduction for them. We also show the relationship between $\sigma$ and $D$ when they are relatively small. Particularly, we have $\sigma \approx D$  when $a/b\to 1$, see SI~Sec.~9 and Fig.~S3. Finally, we derive Eqs.~\eqref{eq:GeneralKuramoto}--\eqref{eq:MemoryDynamics} in the main text, see also SI~Sec.~9.\\
\newline
\noindent{\bf Parameters setting for numerical experiments}\\

\noindent Unless otherwise specified, for numerical simulations, we set the population of oscillators $N=1000$ for the fully-connected networks, $N=200$ for the WS networks with rewiring probability $p=0.1$ and average degree $k=4$. Moreover, we use $N=200$ for both ER and BA networks. The connecting probability for the ER network is $p=0.5$, and the number of node adding each time for generating the BA network is $m=2$. When simulating stochastic differential equations, we use the Euler--Maruyama method, with noise intensity $D=0.5$. The initial phases $\theta_0^{(i)}$ are uniformly distributed on the interval $[0,2\pi]$. The time step is ${\rm d}t=0.05$ and the total simulation time $T=1\times10^4$ to ensure a stable state. The details of the scaling method\cite{Botet1982} for determining the critical coupling strength $\lambda_c$ numerically is demonstrated in SI~Sec.~10. \\
\newline
\noindent{\bf Data and code availability}\\
The authors confirmed that all relevant data are included in the paper and/or its Supplementary Information. Source data are provided with this paper. The source codes for reproducing the {\em neighbor-DTA} and {\em self-DTA} models and the HNN are provided during the reviewing process and they will be publicly available after publication.\\
\newline
\noindent{\bf Acknowledgements}\\
B.-W.Q. is supported by the National Key R\&D Program of China (No.~2024YFA0919500), and by the National Natural Science Foundation of China (NSFC) (No.~12522123 and 12371482). W.L. is supported by the NSFC (No.~11925103~and~12531018), and by the STCSM (Grants No.~2021SHZDZX0103, No.~22JC1402500, No.~22JC1401402~and~No.~25JS2810400), and by the SMEC (Grants No.~2023ZKZD04~and~2023KEJI05-72). This research is funded also by SIMIS under grant number SIMIS-ID-2024-(LZ).
\\
\newline
\noindent{\bf Author contributions}\\
B.-W.Q., K.D. and W.L. conceived the idea, designed the research and directed the project.~Z.Z. performed research.~Z.Z., B.-W. and K.D. analyzed the data.~B.-W.Q. wrote the draft of the paper. All authors revised the paper. \#:~Z.Z. and B.-W.Q. contributed equally to the work.\\
\newline
\noindent{\bf Competing Interests}\\
The authors declare no competing interests.\\
\newline
\Letter~{\bf Corresponding authors}\\
\phantom{\Letter}~boweiqin@fudan.edu.cn;\\
\phantom{\Letter}~kdu@fudan.edu.cn;\\
\phantom{\Letter}~wlin@fudan.edu.cn\\
\newline

\makeatletter
\renewcommand*{\fnum@figure}{{\normalfont\bfseries Extended Data Fig.~\thefigure}}
\renewcommand*{\@caption@fignum@sep}{\textbf{\,$|$\,}}
\makeatother
\makeatletter
\renewcommand*{\fnum@table}{{\normalfont\bfseries Extended Data Table~\thetable}}
\renewcommand\thetable{\arabic{table}}
\makeatother
\makeatletter
\@fpsep\textheight
\makeatother
\stepcounter{myfigure}
\newpage
\begin{figure*}[htb]
	\includegraphics{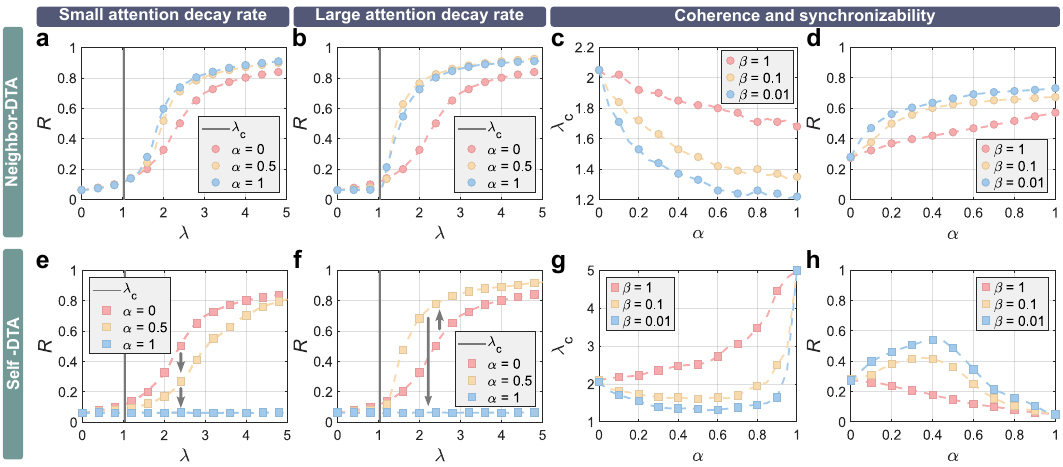}
	\caption{\label{ExFig1} {\bf The effect of DTA in the WS networks.} Panels {\bf a}--{\bf d} and {\bf e}--{\bf h} show the results after applying the {\it neighbor-} and {\it self-DTA} mechanisms. In all cases, the natural frequency distribution $g(\omega)$ is set as the normal distribution $\mathcal{N}(0,0.01)$. {\bf a}, Order parameter $R$ for different coupling strength $\lambda$ under different attention weight (red: $\alpha=0$; yellow: $\alpha=0.5$; blue: $\alpha=1$) when attention decay rate $\beta=1$. {\bf b}, Order parameter $R$ versus $\lambda$ with the same settings as those in panel {\bf a} except that $\beta=0.01$. {\bf c}, Critical coupling strength $\lambda_c$ for different $\alpha$ under distinct $\beta$ (red: 1; yellow: 0.1; blue: 0.01). {\bf d}, Order parameter $R$ for different $\alpha$ with fixed $\lambda=1.5$. The results indicate that the {\it neighbor-DTA} promotes the coherence and synchronizability monotonically with respect to the attention weight $\alpha$. {\bf e}, Order parameter $R$ for different coupling strength $\lambda$ when $\beta=1$. {\bf f}, Order parameter $R$ versus $\lambda$ when $\beta=0.01$. {\bf g}, Critical coupling strength $\lambda_c$ for different $\alpha$. Non-monotonicity appears for small $beta$ (yellow and blue). {\bf h}, Order parameter $R$ for different $\alpha$ with fixed $\lambda=1.5$. The non-monotonicity is consistent with that in panel {\bf g}. The results suggest that the {\it self-DTA} suppresses the coherence and synchronizability monotonically with respect to the increasing of $\alpha$ when the attention decay rate is large, i.e., $\beta=1$. When $\beta=0.1$ and $1$, non-monotonic results appear.}
\end{figure*}
\newpage
\begin{figure*}[!t]
	\includegraphics{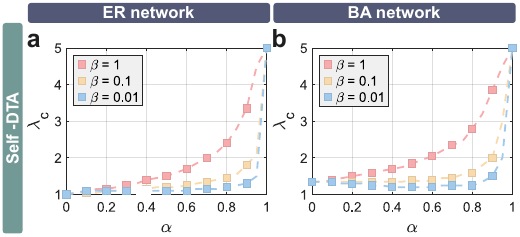}
	\caption{\label{ExFig2} {\bf The effect of} {\it \textbf{self-DTA}}{\bf.} Panels {\bf a} and {\bf b} show the critical coupling strength $\lambda_c$ for different attention weight $\alpha$ with distinct decay rate $\beta$ (red: 1; yellow: 0.1; blue: 0.01) in different networks. The natural frequency distribution is set as $g(\omega)=\delta(0)$ for both panels. {\bf a}, The ER network with connection probability $p=0.5$. {\bf b}, The BA network with 2 individuals added each time when constructing the network.}
\end{figure*}
\newpage
\begin{table*}[t]
	\renewcommand\arraystretch{1.2}
	\setlength\aboverulesep{0pt}
	\setlength\belowrulesep{0pt}
	\centering{\begin{tabular}{cccccc}
			\toprule[2pt]
			\multicolumn{2}{c}{{\bf Spatial topology}} & \multicolumn{2}{c}{{\bf ~~Average shortest path length (ASPL)~~}} & \multicolumn{2}{c}{{\bf The role of {\em self-DTA}}}\\
			\midrule[1pt]
			\rowcolor{gray!10}[2pt][2pt]&&&&&\\
			\rowcolor{gray!10}[2pt][2pt]
			\multicolumn{2}{c}{\multirow{-2}{*}{~~~Fully-connected network~~~}} & \multicolumn{2}{c}{\multirow{-2}{*}{1}} & \multicolumn{2}{c}{\multirow{-2}{*}{Monotonic}} \\
			\rowcolor{gray!0}[2pt][2pt]&&&&&\\
			\rowcolor{gray!0}[2pt][2pt]
			\multicolumn{2}{c}{\multirow{-2}{*}{ER network}}& \multicolumn{2}{c}{\multirow{-2}{*}{1.51}} & \multicolumn{2}{c}{\multirow{-2}{*}{Monotonic}}  \\
			\rowcolor{gray!10}[2pt][2pt]&&&&&\\
			\rowcolor{gray!10}[2pt][2pt]
			\multicolumn{2}{c}{\multirow{-2}{*}{PT network}} & \multicolumn{2}{c}{\multirow{-2}{*}{2.53}} & \multicolumn{2}{c}{\multirow{-2}{*}{Monotonic}} \\
			\rowcolor{gray!0}[2pt][2pt]&&&&&\\
			\rowcolor{gray!0}[2pt][2pt]
			\multicolumn{2}{c}{\multirow{-2}{*}{{\bf BA network}}} & \multicolumn{2}{c}{\multirow{-2}{*}{{\bf 3.26}}} & \multicolumn{2}{c}{\multirow{-2}{*}{{\bf Non-monotonic}}} \\
			\rowcolor{gray!10}[2pt][2pt]&&&&&\\
			\rowcolor{gray!10}[2pt][2pt]
			\multicolumn{2}{c}{\multirow{-2}{*}{{\bf Food network}}} & \multicolumn{2}{c}{\multirow{-2}{*}{{\bf5.09}}} & \multicolumn{2}{c}{\multirow{-2}{*}{{\bf Non-monotonic}}} \\
			\rowcolor{gray!0}[2pt][2pt]&&&&&\\
			\rowcolor{gray!0}[2pt][2pt]
			\multicolumn{2}{c}{\multirow{-2}{*}{{\bf WS network}}} & \multicolumn{2}{c}{\multirow{-2}{*}{{\bf5.95}}} & \multicolumn{2}{c}{\multirow{-2}{*}{{\bf Non-monotonic}}} \\
			\bottomrule[2pt]
	\end{tabular}} 
	\caption{\label{ExTab1} {\bf The relationship between ASPL and the role of {\em self-DTA}.} Networks with different spatial topologies and their ASPL are summarized.~For the spatial networks with large ASPL, the {\it self-DTA} exhibits a non-monotonic effect on promoting synchronizability and coherence. Particularly, there is an optimal attention weight $\alpha$ for promoting synchronization.}
\end{table*}
\end{document}